\documentclass[conference]{IEEEtran}
\IEEEoverridecommandlockouts
\usepackage{cite}
\usepackage{amsmath,amssymb,amsfonts}
\usepackage{algorithm}
\usepackage{graphicx}
\usepackage{textcomp}
\usepackage{csquotes}
\usepackage{multirow}
\usepackage{algpseudocode}
\usepackage{xcolor}
\usepackage[top=0.75in, left=0.75in, right=0.75in, bottom=1.1in]{geometry}
\def\BibTeX{{\rm B\kern-.05em{\sc i\kern-.025em b}\kern-.08em
    T\kern-.1667em\lower.7ex\hbox{E}\kern-.125emX}}
\begin{document}

\title{Optimizing Search and Rescue UAV Connectivity in Challenging Terrain through Multi Q-Learning}

\author{\IEEEauthorblockN{Mohammed M. H. Qazzaz$^{1,2}$\qquad Syed A. R. Zaidi$^{1}$ \qquad Desmond C. McLernon$^{1}$ \\ \qquad Abdelaziz Salama$^{1}$ \qquad  Aubida A. Al-Hameed$^{2}$}
\IEEEauthorblockA{\textit{$^{1}$ School of Electronic and Electrical Engineering,}
\textit{University of Leeds,} Leeds, UK \\
\textit{$^{2}$ College of Electronics Engineering,}
\textit{Ninevah University,} Mosul, Iraq \\
Corresponding Author: Mohammed M. H. Qazzaz (ml14mmh@leeds.ac.uk)}
}

% \IEEEoverridecommandlockouts
% \IEEEpubid{\makebox[\columnwidth]{979-8-3503-7786-6/24/\$31.00~\copyright~2024~IEEE \hfill} \hspace{\columnsep}\makebox[\columnwidth]{ }}

\maketitle

\IEEEpubidadjcol

\begin{abstract}
Using Unmanned Aerial Vehicles (UAVs) in Search and rescue operations (SAR) to navigate challenging terrain while maintaining reliable communication with the cellular network is a promising approach. This paper suggests a novel technique employing a reinforcement learning multi Q-learning algorithm to optimize UAV connectivity in such scenarios. We introduce a Strategic Planning Agent for efficient path planning and collision awareness and a Real-time Adaptive Agent to maintain optimal connection with the cellular base station. The agents trained in a simulated environment using multi Q-learning, encouraging them to learn from experience and adjust their decision-making to diverse terrain complexities and communication scenarios. Evaluation results reveal the significance of the approach, highlighting successful navigation in environments with varying obstacle densities and the ability to perform optimal connectivity using different frequency bands. This work paves the way for enhanced UAV autonomy and enhanced communication reliability in search and rescue operations.
\end{abstract}


\section{Introduction}\label{introduction}
\subsection{Motivation}
In recent times, unmanned aerial vehicles (UAVs) have garnered significant interest due to their versatility across civilian and military sectors. Capitalizing on their rapid mobility, autonomous functionality, adaptable deployment, and cost-effectiveness, UAVs find applications in various fields, including but not limited to final emergency response, parcel delivery, medical aid, and maintaining wireless network coverage. Additionally, advancements in Edge computing have made it possible to integrate Artificial Intelligence (AI) and Machine Learning (ML) algorithms, such as Reinforcement Learning (RL) tasks, along with visualization engines to facilitate Virtual Reality (VR) support directly onboard UAVs.\\

In the domain of Search and Rescue (SAR) operations, the deployment of UAVs has become increasingly vital. These UAVs offer unparalleled advantages in navigating challenging terrain, enabling swift response and efficient search capabilities. However, ensuring robust connectivity for SAR UAVs, especially in remote and rugged environments, presents significant challenges.

Traditional approaches to SAR UAV connectivity often fail in challenging terrain. Satellite channels, while offering coverage, suffer from high latency and prohibitive costs. Dedicated UAV-specific channels entail substantial infrastructure investment, making them impractical for widespread deployment. Integration with cellular networks presents a promising solution, leveraging existing infrastructure and advancements in wireless technology.

Despite the potential of cellular connectivity, optimizing SAR UAV trajectories in challenging terrain remains a critical yet unresolved issue. Standard trajectory models fail to account for the dynamic and unpredictable nature of SAR missions, where obstacles and environmental conditions constantly evolve. Ensuring continuous connectivity while navigating complex landscapes requires a realistic approach that can adapt in real-time.

In response to these challenges, this paper proposes a pioneering framework for optimizing connectivity for SAR UAVs in challenging terrain through Multi Q-Learning algorithms. By harnessing the power of reinforcement learning, our approach aims to dynamically adjust UAV trajectories to maintain seamless connectivity with ground stations while navigating through rugged landscapes.

This novel methodology not only addresses the unique challenges of SAR operations but also paves the way for more efficient and effective search and rescue missions. By integrating connectivity optimization with trajectory planning, we strive to enhance the capabilities of SAR UAVs, ultimately saving lives and mitigating the impact of disasters in even the most challenging terrains.

\subsection{Related Works}
In recent years, the utilization of UAVs for SAR operations has garnered significant attention owing to their potential to enhance efficiency and safety in emergency situations \cite{san2018intelligent,qazzaz2024non,salama2023flcc}. Several methods have been proposed for optimizing UAV path planning in SAR missions, aiming to minimize risk exposure while maximizing coverage \cite{san2018intelligent}. These methods often leverage intelligent algorithms, such as fuzzy logic, particle swarm optimization (PSO), and genetic algorithms, to calculate optimal UAV trajectories and waypoints \cite{oh2021smart,zhang2023novel}. Additionally, efforts have been made to address collision avoidance challenges in UAV swarm operations, with approaches like multi-plane systems and collision avoidance algorithms demonstrating promising results \cite{kumar2022obstacle,du2023multi,qazzaz2023low}.

Communication plays a crucial role in multi-UAV path planning for SAR missions, enabling dynamic task allocation and information dissemination \cite{hayat2020multi}. Strategies such as simultaneous inform and connect (SIC) path planning have been proposed to optimize mission tasks while maintaining connectivity and coverage goals \cite{hayat2020multi}. Furthermore, advancements in Fifth Generation (5G) mobile networks offer opportunities to enhance SAR missions through dynamic and autonomous placement of Network Functions (NFs) and Artificial Intelligence (AI) systems \cite{9508372,9836167}. These developments aim to leverage edge intelligence and system intelligence concepts to optimize UAV-based SAR operations \cite{9508372,qazzaz2024machine}.

Efficient path planning algorithms are essential for optimizing UAV trajectories in SAR missions. Techniques such as A algorithm and task allocation algorithms have been enhanced to achieve faster and more effective path planning, particularly in cluttered environments \cite{du2023multi,8794345}. Moreover, collaborative efforts in the research community have led to the development of multi-robot systems supporting SAR operations, focusing on algorithmic perspectives for multi-robot coordination and perception \cite{9220149}. These systems aim to address various challenges, including shared autonomy, sim-to-real transferability, and active perception \cite{9220149}.\\

\subsection{Contributions and Problem Statement}
In this paper, we tackle the pivotal challenge of enhancing connectivity for SAR UAVs operating in challenging terrains. The motivation behind our work arises from the imperative need to ensure seamless communication between SAR UAVs and ground base stations, particularly in environments characterized by rugged landscapes and adverse conditions.

Our study's primary contribution lies in proposing a novel approach that leverages multi Q-learning algorithms to optimize connectivity for SAR UAVs in challenging terrains while achieving the shortest travel time. SAR missions often unfold in remote or inaccessible areas, where direct communication with the UAV operator may be inadequate or non-existent. Consequently, SAR UAVs face significant hurdles in maintaining continuous connectivity with ground base stations, risking mission effectiveness and the safety of both rescuers and victims.

Our approach addresses this critical issue by integrating multi Q-learning algorithms into UAV trajectory planning. We prioritize optimal paths that ensure uninterrupted connectivity while navigating challenging terrain. By harnessing the power of reinforcement learning, our methodology aims to dynamically adapt UAV trajectories in real-time, accounting for environmental obstacles, terrain variations, and communication dynamics.

\subsection{Organisation}
This paper is structured as follows: Section \ref{RL_sec} provides an overview of the Reinforcement Learning (RL) model, explaining its fundamental architecture and mathematical formulation. Subsequently, Section \ref{sys_model} outlines the proposed model, encompassing the employed environment and implemented policies. The outcomes of the trained model and relevant testing observations are detailed in Section \ref{results_sec}. Lastly, Section \ref{concl} encapsulates our conclusions drawn from the study.

%%%%%%%%%%  Section II - Reinforcement Learning Model %%%%%%%%%%%%%
\section{Reinforcement Learning (RL) Algorithm}\label{RL_sec}
One of the current thrilling areas of study in AI/ML is RL. In this paper, our focus lies on implementing RL, where a dynamic \enquote{agent} navigates through various possible circumstances and makes decisions within an operational \enquote{environment} to ultimately determine the optimal ones. In RL, every action taken by the agent is denoted as an \enquote{action}, and each action leads to a subsequent state. These transitions are evaluated to determine their effectiveness, with successful transitions being considered for future iterations to reinforce positive behaviour \enquote{rewards}, while unsuccessful transitions are neglected to avoid repeating the poor performance \enquote{penalties}. Through this iterative process, the agent learns to distinguish between positive and negative actions, thus anticipating successful actions in real-world scenarios by adjusting the rewards and penalties accordingly \cite{sutton2018reinforcement}.

\begin{figure}
  \includegraphics[width=3.45 in]{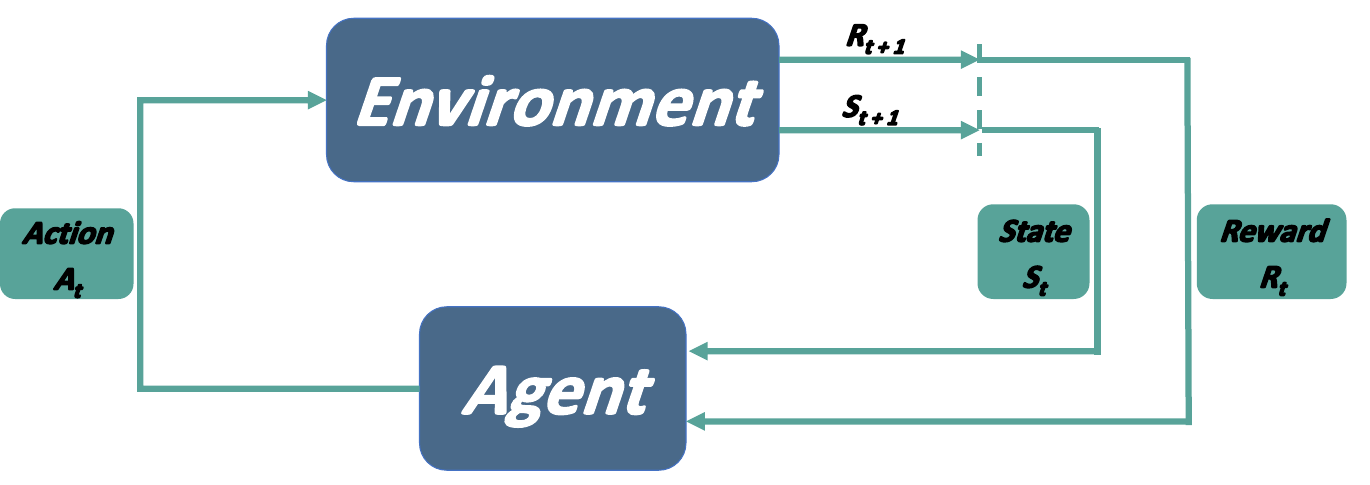}
 % \vspace{-15pt}
  \caption{Reinforcement Learning Model \cite{munoz2019deep}}\label{RL}
\end{figure}

Figure \ref{RL} illustrates the fundamental architecture of RL, depicting the iterative process of action and state transitions.

Q-learning stands out as one of the most widely adopted reinforcement learning algorithms. Within this framework, the agent endeavours to select the optimal action by leveraging information from past states and associated rewards. Each iteration of this process termed an \enquote{episode}, contributes to accumulating knowledge about optimal actions and rewards.
Following each episode, the rewards or penalties earned for each action and state are stored in a lookup table known as the Q-table. This table serves as a valuable resource for the agent, guiding its decision-making process by informing which actions to take to maximize rewards and avoid penalties \cite{sutton2018reinforcement}.
\begin{equation}
\begin{split}
Q_{t+1}(S_t, A_t) \leftarrow Q(S_t, A_t) + \alpha[ R_{t+1} + \gamma \max_a Q(S_{t+1}, A_{t}) \\ 
- Q(S_t, A_t)]
\end{split}
\label{QL_Eq}
\end{equation}

The equation \ref{QL_Eq} provides the formal definition of the Q-learning algorithm, where the variables represent the following: $S_t$: The state of the environment at time $t$, $A_t$: The action taken by the system at time $t$ to change the state to $S_{t+1}$, $R_{t+1}$: The reward received after taking action $A_t$ at time $t$, $\alpha$: The learning rate, determining the extent to which new information overrides old information and $\gamma$: The discount factor, representing the importance of future rewards relative to immediate rewards.

\section{System Model}\label{sys_model}
This study focuses on optimizing the connectivity and path planning for search and rescue UAVs. The application scenario involves deploying UAVs to conduct search and rescue missions in remote and difficult-to-access areas.
Depending on flight conditions, the UAV may necessitate traversing vast distances and dealing with tough terrains, sometimes visiting areas inaccessible to human intervention. Consequently, such missions demand careful path planning to mitigate the risk of crashes and ensure seamless connectivity with the remote controller or pilot. Our proposed approach involves integrating an autonomous model onto the UAV, facilitating its adept navigation between the originating and target locations, thereby optimizing operational efficiency.

\subsection{Model Environement}\label{model_env}

\begin{figure}\
\centering
  \includegraphics[width=3.05 in]{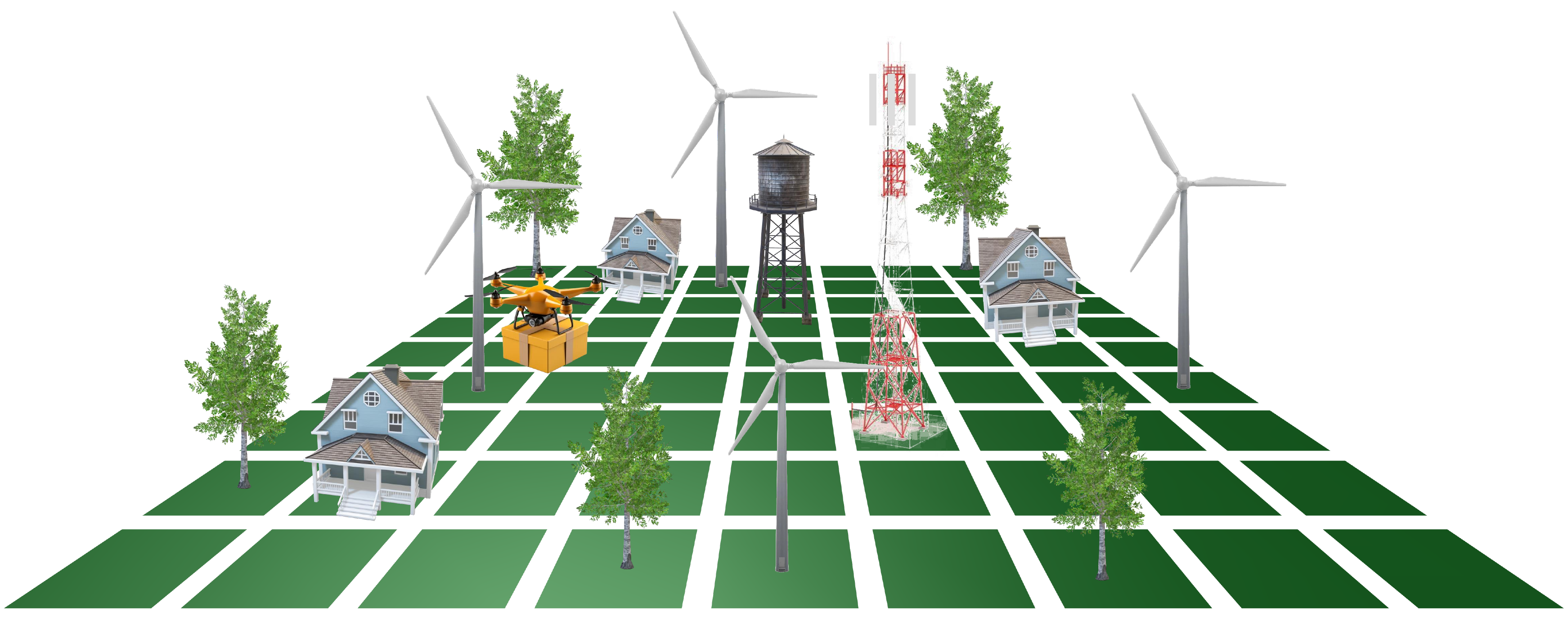}
 % \vspace{-15pt}
  \caption{Supposed Model 3D Environement}\label{env}
\end{figure}

A 3D environment has been built on a grid-like depiction of the region to employ this model, as shown in Figure~\ref{env}, to develop and test reinforcement learning on an agent with the parameters listed in Table~\ref{Env_para}. This environment represents the rugged terrain and potential obstacles that UAVs may encounter during operations. Our approach is intended to autonomously operate the UAV, which connects to and is controlled by a ground base station inside a specific region through cellular service. The RL agent target must adhere to the coverage channel constraints to keep connected to the base station and the operator, avoid any on-way obstacles, and take the shortest route between the initial and destination endpoints.

To facilitate the development and evaluation of our model, a three-dimensional (3D) environment, as illustrated in Figure~\ref{env}, has been constructed utilizing a grid-based representation of the designated region. The implementation and validation of reinforcement learning algorithms are conducted within this environment, utilizing an agent configured with parameters delineated in Table~\ref{Env_para}. Our methodology aims to train a UAV which maintains connectivity with a ground cellular base station within a specified geographical area. The reinforcement learning (RL) agent's objective is to adhere to coverage channel constraints to ensure continuous connectivity with both the base station and the operator, navigate around potential obstructions along its flight path, and determine the most efficient path.

To simulate the connectivity between the UAV and the ground base station, we have adopted the COST Hata propagation model. The COST 231 Hata model, also recognized as the Hata-Extended model, is specifically tailored for wireless propagation estimations in rural locations, offering an enhancement over the Okumura-Hata model by integrating terrain roughness and vegetation characteristics, thereby providing more accurate estimations for UAV communication in non-urban regions \cite{dalela2012tuning}. This model characterizes path loss according to the following formulation:

\begin{equation}\label{p_l}
\begin{split}
\mathcal{L}_{b} = 46.3 + 33.9 \log_{10}\frac{f}{\text{MHz}} - 13.82 \log_{10}\frac{h_{B}}{\text{m}} - \alpha \\
+ \left(44.9 - 6.55 \log_{10}\frac{h_{B}}{\text{m}}\right) \log_{10}\frac{d}{\text{km}} + C_{m}
\end{split}
\end{equation}

Here, $f$ denotes the frequency in MHz, $h_B$ signifies the height of the ground base station antenna in metres, $d$ represents the distance between the UAV and the base station antennas in kilometres, and $C_m$ stands for the terrain roughness correction factor. The term $\alpha$ is further delineated as follows:

% \begin{equation}\label{p_l}
% \begin{split}
% \alpha = \left(1.1 \log_{10}{\frac{f}{\text{MHz}}} - 0.7 \right)\frac{h_{R}}{\text{m}} \\
% - \left(1.56 \log_{10}{\frac{f}{\text{MHz}}} - 0.8\right)
% \end{split}
% \end{equation}

\begin{equation}
\alpha = \Bigl(1.1 \log_{10}{\frac{f}{\text{MHz}}} - 0.7 \Bigr)\frac{h_{R}}{\text{m}} - \Bigl(1.56 \log_{10}{\frac{f}{\text{MHz}}} - 0.8\Bigr)
\end{equation}

where $h_R$ corresponds to the height of the UAV in meters.

Various frequency ranges have been explored to ensure the robustness of the trained model across diverse scenarios.

\begin{table}
\centering
\caption{Parameters of Reinforcement Learning Environment}
\begin{center}
\begin{tabular}{ |c| c|  }
\hline
 \textbf{Parameter} & \textbf{Value}  \\ 
 \hline
 Region Area & 1 square km \\  
 \hline
 UAV Max Altitude & 100 m \\
 \hline
 No. of UAVs & 1  \\
 \hline
 UAV Max Velocity & 15 m/s  \\
 \hline
 Base Station Location & ($x_b,y_b$)  \\
 \hline
 Base Station Antennas Height & 60 m \\
 \hline
 Cellular Propagation Model & COST Hata  \\
 \hline
 Frequency Bands & 900, 1800, 2600 MHz  \\
 \hline 
 Learning Rate ${\delta}$ & 0.8 \\
 \hline
 Discount Rate ${\gamma}$ & 0.5 \\
 \hline
\end{tabular}
\end{center}
\label{Env_para}
\end{table}

\subsection{Model Impelementation}
To translate our assumptions into a practical framework, we implemented two agents tailored to the specific requirements of our SAR UAV connectivity optimization model. The implementation involved several policies, each aimed at ensuring the realism and effectiveness of our proposed approach. 

To train the two agents effectively, we employ a combination of exploration and exploitation strategies within the environment, adhering to the constraints provided. The exploration-exploitation trade-off is managed using the ${\epsilon}$-Greedy policy, which guides the agents in selecting their actions during training.

In the ${\epsilon}$-Greedy policy, the agents initially explore the environment by choosing random actions to transition between states, especially in the early stages of training when the Q-values are not well-established. This random exploration allows the agents to gather valuable experience and information about the environment, including potential rewards and penalties associated with different actions.

As training progresses and the Q-values begin to converge, the agents gradually shift towards exploitation, where they prioritize actions with higher estimated Q-values based on their learned experiences. The parameter ${\epsilon}$, known as the greedy rate, governs this transition from exploration to exploitation. ${\epsilon}$ is initially set high to encourage exploration but decreases gradually over time as the agents accumulate more knowledge and update their Q-values.

By dynamically adjusting ${\epsilon}$ based on the progress of training and the reliability of the Q-values, the agents strike a balance between exploring new possibilities and exploiting known strategies. This adaptive approach allows the agents to efficiently learn from their experiences and optimize their actions within the environment, ultimately enhancing their performance and achieving the desired objectives.

\subsubsection{Strategic Planning Agent}
This agent is responsible for non-time-sensitive parameters like determining the shortest path and identifying obstacle locations. We assume that these factors aren't changing much, and there is no need to retrain the agent when other fluctuating factors, like channel connectivity, change. This agent should navigate the environment to find the shortest path between a fixed initial location and random destination points. During the training stage, the agent explores the environment in different positions, taking right and wrong actions to train itself in the right direction. A reward will be received if its step correctly decreases the distance to the destination; otherwise, it will receive a penalty for following the longest path. During its navigation, it checks the location of the obstacles, and once it passes through an obstacle, this means a crash is happened, and a high penalty will be received by the agent.   

The agent responsible for non-time-sensitive parameters, such as determining the shortest path and identifying obstacle locations, operates under the assumption that these factors remain relatively static and do not require frequent retraining. This agent's primary task is to navigate the environment efficiently, finding the shortest path between an initial location and randomly selected destination points as shown in Algorithm \ref{alg_static}.

During the training stage, the agent systematically explores the environment, making decisions to advance towards the destination. It learns from both successful and unsuccessful actions, receiving rewards when its actions effectively reduce the distance to the destination and penalties when it deviates from the optimal path.

As the agent navigates the environment, it continuously evaluates the presence and location of obstacles. Upon encountering an obstacle, signifying a potential collision, the agent incurs a substantial penalty. This penalty serves as a reinforcement mechanism, discouraging the agent from choosing actions that lead to collisions and encouraging it to prioritize obstacle-free paths.

The agent gradually refines its navigation strategy by iteratively exploring the environment, learning from experiences, and adapting its behaviour based on received rewards and penalties. This process develops a robust understanding of the environment's layout, identifying optimal paths while avoiding obstacles to effectively fulfil its designated tasks.

\begin{algorithm}

\caption{Strategic Planning Agent Training Algorithm}\label{alg_static}
 
\begin{algorithmic}

\State \textbf{Initialize:} Hyper-parameters: learning rate $\alpha \in \{0,1\}$,~\textnormal{discount rate}~$\gamma \in \{0,1\}$,~\textnormal{greedy rate}~$\epsilon \in \{0,1\}$

\State \textbf{Initialize:} (i) Environment Dimensions, (ii) Initial coordinate, (iii) No. of Episodes ($M$), (iv) Density of Obstacles

\State \textbf{Initialize:} action-value function $Q$, 

\While{Episode $<$  No. of Episodes}
    \State \textbf{Initialize:} Destination coordinate (Dest.)
    
    \State \textbf{Reset} Current State $S_{t}$ to initial location

    \State \textbf{Reset} Rewards to 0

    \State \textbf{Calculate} $Distance_{t}$ between $S_{t}$ and Dest.

    \While{Destination not arrived}

        \State \textbf{Choose} action $A_t$ according to:
                \[ A_t = \begin{cases} 
                \text{random action} & \text{with probability } \epsilon \\
                \arg\max_a Q(S_t, a) & \text{with probability } 1 - \epsilon 
                \end{cases}
                \]

        \State \textbf{Select} Next State $S_{t+1}$ based on $A_t$

        \State \textbf{Calculate} $Distance_{t+1}$ between $S_{t+1}$ and Dest.

        \If{$Distance_{t+1}$ $<$ $Distance_{t}$}

            \State The agent gets a reward

        \Else

            \State The agent gets a penalty

        \EndIf 

        \If{$S_{t+1}$ == Obstacle location}

            \State The agent gets a high penalty

        \EndIf 

        \If{$S_{t+1}$ == Destination}

            \State The agent gets a high reward

        \EndIf 

        \State \textbf{Update} Q-value for the current state-action pair

    \EndWhile

\EndWhile

\end{algorithmic}

\end{algorithm}

\subsubsection{Real-time Adaptive Agent}
The Real-time Adaptive Agent is tasked with identifying optimal coverage locations within the environment to ensure continuous connectivity between the UAV and the base station during flight operations. Unlike non-time-sensitive parameters, such as path planning and obstacle avoidance, connectivity conditions can vary dynamically over time. Therefore, this agent operates independently to adapt to changing connectivity requirements, distinct from the Strategic Planning Agent.

The primary objective of the Real-time Adaptive Agent is to identify and navigate through regions with optimal coverage to maintain seamless communication with the base station. This entails strategically selecting flight paths that maximize signal strength and minimize the risk of signal loss or degradation.

During the training phase, the agent explores the environment systematically, evaluating coverage requirements at different locations and frequency bands. As shown in Algorithm \ref{alg_dynamic}, this agent receives rewards for crossing spots with accepted coverage, ensuring reliable communication with the base station. In contrast, penalties are incurred for steering through areas with insufficient coverage, leading to possible communication troubles.

Multiple frequency bands, including 900, 1800, and 2100 MHz, have been considered with the Real-time Adaptive Agent to sustain the variability in connectivity conditions. Training this agent across different frequency bands makes it capable of adapting to various operating scenarios, ensuring compatibility with various communication circumstances.

The Real-time Adaptive Agent's capability to assess and adapt to changing connectivity requirements improves the overall robustness 
and reliability of the cellular connected UAV. By prioritizing optimal coverage locations with different frequency bands, the agent plays an essential role in maintaining continuous connectivity, thereby enabling efficient and effective UAV operations in challenging atmospheres.

\begin{algorithm}

\caption{Real-time Adaptive Agent Training Algorithm}\label{alg_dynamic}
 
\begin{algorithmic}

\State \textbf{Initialize:} Hyper-parameters: learning rate $\alpha \in \{0,1\}$,~\textnormal{discount rate}~$\gamma \in \{0,1\}$,~\textnormal{greedy rate}~$\epsilon \in \{0,1\}$

\State \textbf{Initialize:} (i) Environment Dimensions, (ii) base station coordinate, (iii) No. of Episodes ($M$), (iv) frequency range, (v) $SNR$ threshold

\State \textbf{Initialize:} action-value function $Q$, 

\While{Episode $<$  No. of Episodes}
    \State \textbf{Initialize:} Destination coordinate (Dest.)
    
    \State \textbf{Reset} Current State $S_{t}$ to initial location

    \State \textbf{Reset} Rewards to 0
    
    \While{Destination not arrived}

        \State \textbf{Choose} action $A_t$ according to:
                \[ A_t = \begin{cases} 
                \text{random action} & \text{with probability } \epsilon \\
                \arg\max_a Q(S_t, a) & \text{with probability } 1 - \epsilon 
                \end{cases}
                \]

        \State \textbf{Select} Next State $S_{t+1}$ based on $A_t$

        \State \textbf{Calculate} $\mathcal{L}_{b}$ and $SNR$ of $S_{t+1}$

        \If{$SNR$ $<$ $SNR$ threshold}

            \State The agent gets a penalty

        \Else

            \State The agent gets a reward

        \EndIf 
        
        \State \textbf{Update} Q-value for the current state-action pair

    \EndWhile

\EndWhile

\end{algorithmic}

\end{algorithm}

\subsubsection{Dual Model Decision-Making Approach}
A dual-agent decision-making approach is implemented to leverage the capabilities of both the Strategic Planning Agent and the Real-time Adaptive Agent effectively. This approach aims to select the most beneficial action for the UAV while considering inputs from both agents and their respective Q-value tables.

Both agents experience comprehensive offline training to learn optimal strategies for their assigned missions. During this stage, the Strategic Planning Agent focuses on path planning and obstacle avoidance. Simultaneously, the Real-time Adaptive Agent concentrates on identifying optimal coverage locations to maintain connectivity with the ground base station. Additionally, the Real-time Adaptive Agent might undergo online training to mitigate fluctuations in channel conditions to ensure continuous connectivity during the UAV operations.

Algorithm \ref{alg_decision} outlines the decision-making process for the dual-agent approach. At each time step $t$, both agents independently select their best actions $A_1$ and $A_2$ based on their respective Q-values for the current state $S_t$. The UAV then determines the best course of action based on the outcomes of both agents' selections.

\begin{figure*}
\centering
  \includegraphics[width=7. in]{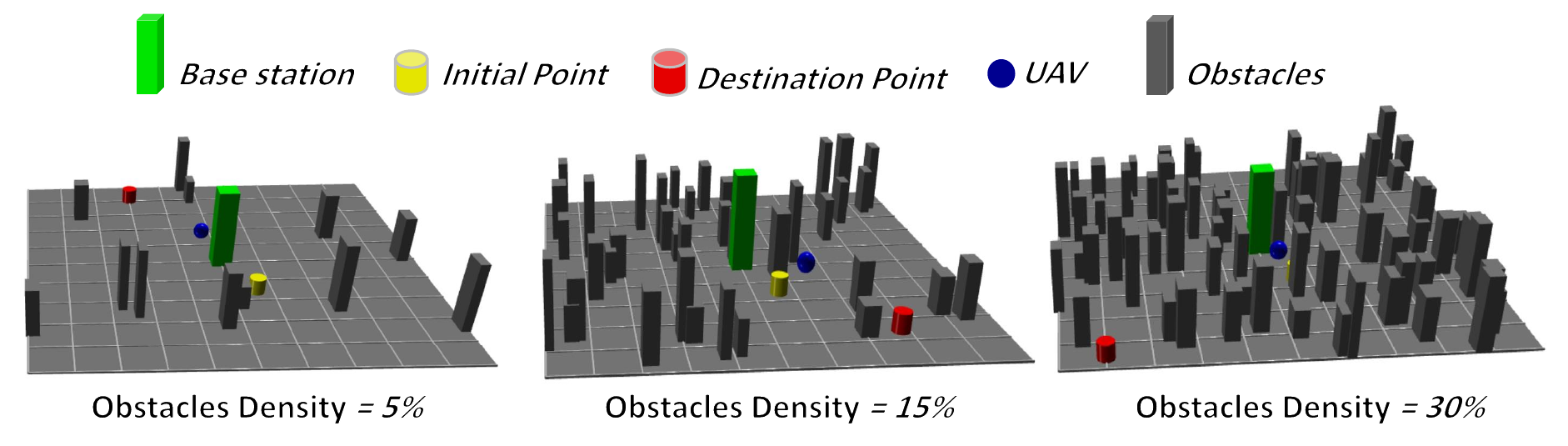}
 % \vspace{-15pt}
  \caption{Environement Obstacles Distribution Scenarios}\label{env_obs}
\end{figure*}

In scenarios where the Strategic Planning Agent and the Real-time Adaptive Agent choose different actions, a hierarchical approach is employed to prioritize decision-making. The importance of avoiding obstacles takes precedence over navigating through areas with weak coverage. Therefore, the UAV prioritizes the action recommended by the Strategic Planning Agent when conflicts arise. In certain situations, the optimal path recommended by the Strategic Planning Agent may lead through areas with weak coverage. In such cases, the UAV may opt to take an additional step, resulting in a longer path to the destination, to ensure continuous connectivity with the ground base station.

This approach ensures that the UAV optimally balances path optimization and connectivity considerations, resulting in enhanced performance and reliability in challenging environments.

\begin{algorithm}

\caption{Decision-Making Algorithm}\label{alg_decision}
 
\begin{algorithmic}

\State \textbf{Specify:} Current State $S_{t}$

\State \textbf{Choose} action $A_1$ according to Strategic Planning Agent

\State \textbf{Choose} action $A_2$ according to Real-time Adaptive Agent

\If{$A_1$ $=$ $A_2$}

    \State \textbf{Select} Next State $S_{t+1}$ based on $A_1$

\Else

    \State \textbf{Check} The $Q$ value from the Strategic Planning Agent $Q_1$ for $A_2$

    \State \textbf{Check} The $Q$ value from the Real-time Adaptive Agent $Q_2$ for $A_1$ 

    \If{$Q_1$ $>$ $Q_2$}
        State \textbf{Select} Next State $S_{t+1}$ based on $A_2$

    \Else
        \State \textbf{Select} Next State $S_{t+1}$ based on $A_1$

    \EndIf

\EndIf

\end{algorithmic}

\end{algorithm}

\section{Results and Discussion}\label{results_sec}
In this section, we assess the performance of our proposed models through extensive training and evaluation processes. The evaluation focuses on analyzing the training progress and the effectiveness of the trained agents in navigating the environment and optimizing UAV operations. 

During each agent's initial training episodes, the Q-model lacks familiarity with the environment states, leading to significant penalties due to incomplete information. Consequently, more training episodes may be required to explore the vast state space effectively and optimize decision-making processes.

The training of the Strategic Planning Agent involves navigating the environment while considering several densities of obstacles, as shown in Figure \ref{env_obs}, and determining the shortest path to a random destination. At each step, the agent calculates the distance to the destination, evaluates the possibility of a crash, and receives rewards or penalties accordingly. These experiences are used to update the Q-values, guiding the agent towards optimal decision-making.

Similarly, the Real-time Adaptive Agent undergoes training to assess the connection performance between the UAV and the ground base station using three different frequency bands. Rewards and penalties are assigned based on the agent's decisions, considering factors such as the received signal strength and signal-to-noise ratio. The Q-values are updated iteratively, enabling the agent to learn from past actions and improve its decision-making capabilities.

After multiple training trials, both agents learn from their experiences and begin to select better actions to maximize rewards and minimize penalties. The average rewards and penalties accumulated during the training phase are used as performance metrics to gauge the agents' effectiveness.

\begin{figure}\
\centering
  \includegraphics[width=3.25 in]{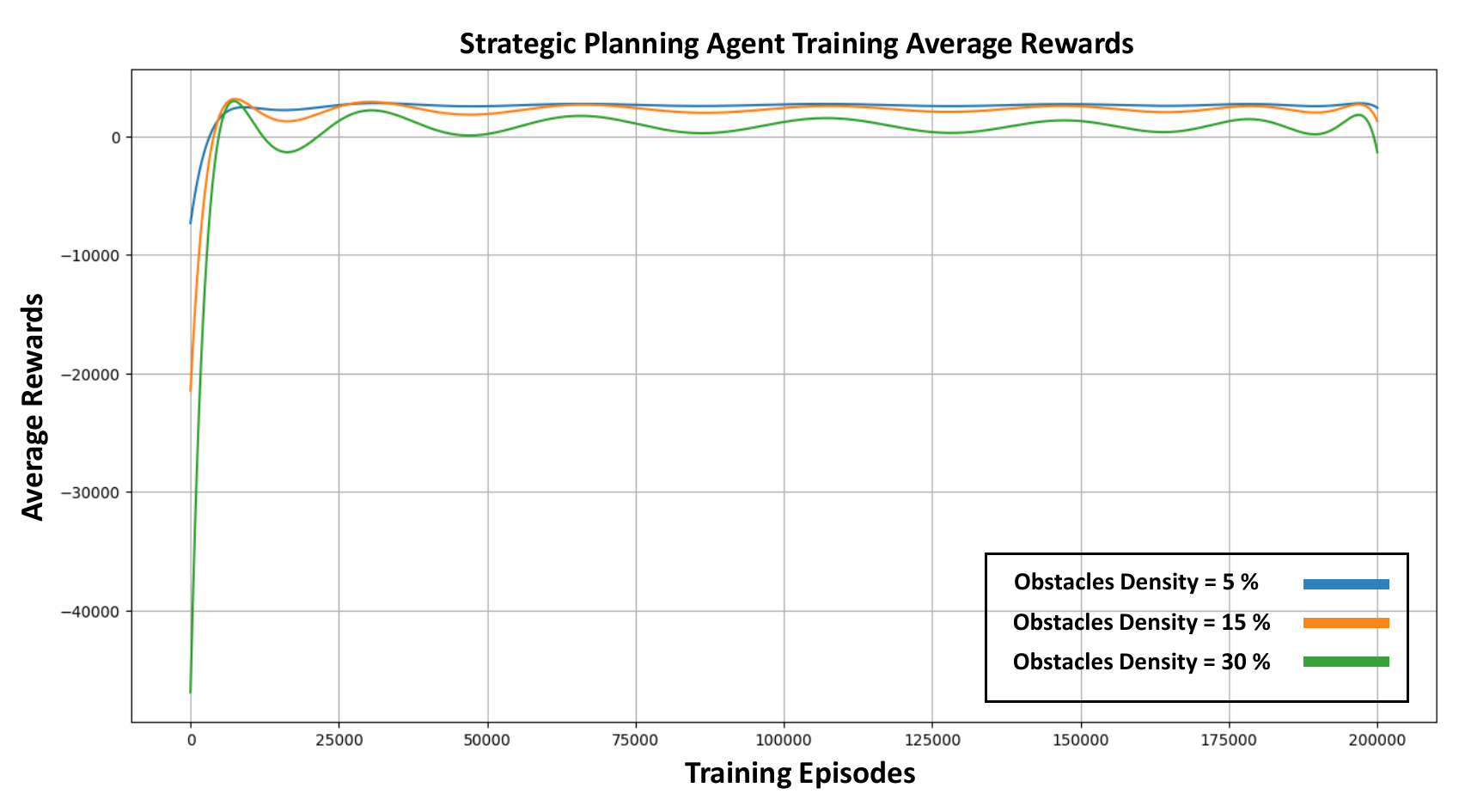}
 % \vspace{-15pt}
  \caption{Strategic Planning Agent Training Rewards}\label{reward_1}
\end{figure}

Figure \ref{reward_1} illustrates the average rewards and penalties obtained by the Strategic Planning Agent over multiple training trials. It is evident that the agent learns its optimal actions more quickly in environments with fewer obstacles, as indicated by the rapid transition to positive rewards. However, in environments with a higher density of obstacles, the agent faces greater challenges in identifying optimal actions and avoiding collisions. This observation underscores the importance of the environment complexity in shaping the learning dynamics of the agent.

\begin{table*}[t]
\centering
\caption{Trained Agents Evaluation Results}
\begin{tabular}{|c|p{3.5cm}|p{3.6cm}|p{3.6cm}|} 
\hline
\multirow{2}{*}{\textbf{Environment Testing Conditions}} & \textbf{Frequency Band = 900 MHz, Obstacles Density = 5\%} & \textbf{Frequency Band= 1800 MHz, Obstacles Density = 15\%} & \textbf{Frequency Band = 2100 MHz, Obstacles Density = 30\%} \\ 
\hline
Arrival Success   & 100\%  & 100\%   & 90\% \\ 
\hline
Obstacles Crash Events & 0\%  & 0\%   & 10\% \\ 
\hline
Outage Events & 1\% & 10\%  & 25\%  \\ 
\hline
\end{tabular}
\label{Eva}
\end{table*}

\begin{figure}
\centering
  \includegraphics[width=3.25 in]{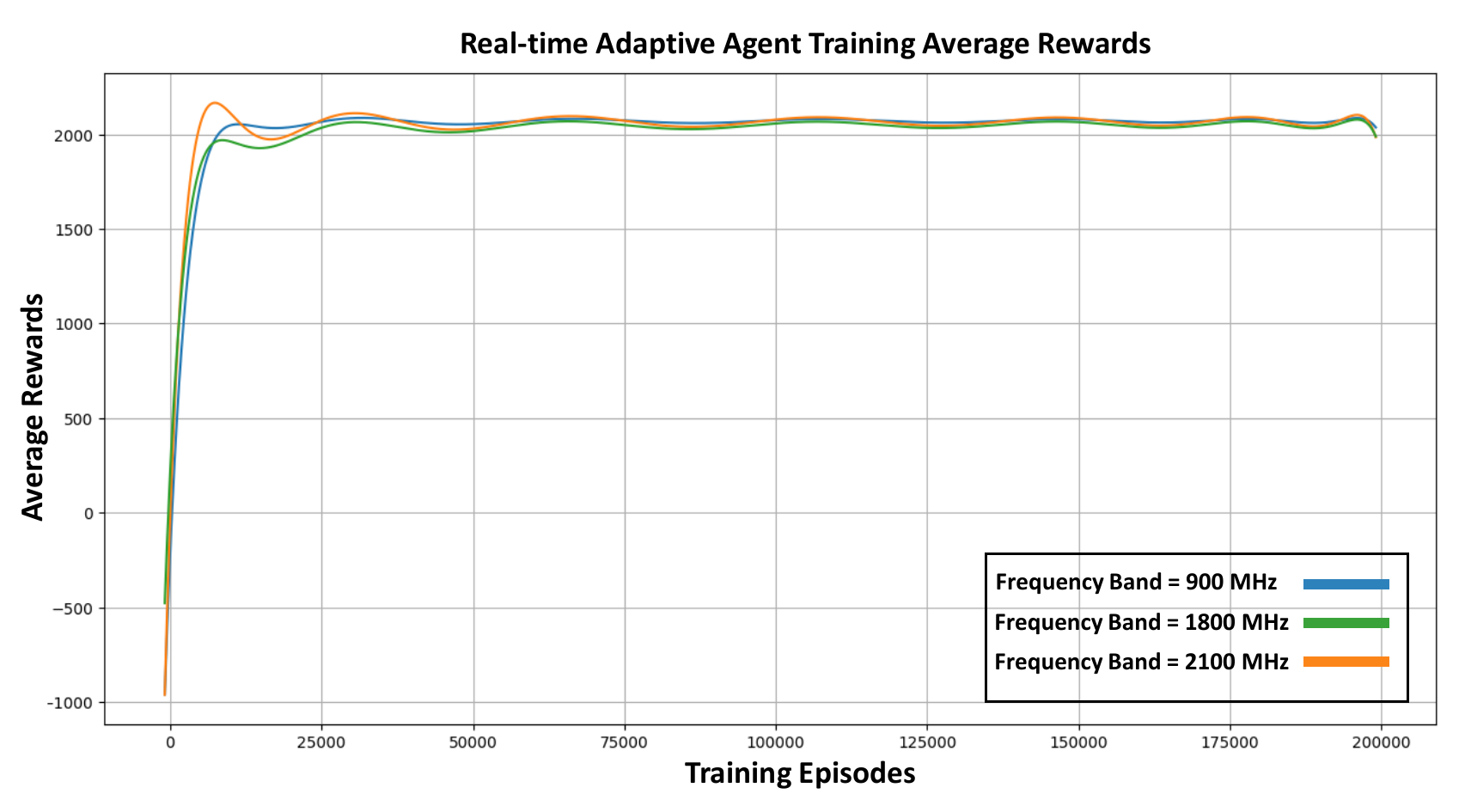}
 % \vspace{-15pt}
  \caption{Real-time Adaptive Agent Training Rewards}\label{reward_2}
\end{figure}

In contrast, Figure \ref{reward_2} presents the corresponding metrics for the Real-time Adaptive Agent. These figures offer valuable insights into the training dynamics and the agents' learning progress over time. Notably, the impact of changing frequency bands on the agent's performance is apparent in the first training stages, when the agent faced a lot of weak coverage locations. As expected, higher frequency bands, such as 2100 MHz, result in increased path loss due to factors like scattering and absorption by vegetation and atmospheric gases. Consequently, the agent encounters difficulty in identifying optimal connectivity points in environments characterized by higher frequency bands, leading to prolonged learning periods and higher penalties. Conversely, lower frequency bands, such as 900 MHz, facilitate faster learning by providing more favourable connectivity conditions and enabling the agent to identify optimal actions more efficiently.

These observations highlight the significance of environmental factors and frequency band selection in shaping the learning dynamics and performance of the Real-time Adaptive Agent. By understanding these dynamics, we can optimize training strategies and enhance the agent's ability to navigate diverse environments effectively.

To assess the performance of our trained agents, we conducted tests in environments mirroring the dimensions, UAV taking-off location, base station location, and connectivity settings used during training. Table \ref{Eva} presents an overview of our trained system's accuracy across different scenarios.

Notably, the first two scenarios achieved a 100\% success rate in terms of reaching the destination and successfully avoiding on-path obstacles during flight. However, during the evaluation stage, where the focus shifted to successful delivery more than navigating through weak coverage points, UAV agents experienced outage events in approximately 1\% of flights in the low-frequency band and 10\% in the middle-frequency band. These events highlight the inherent trade-off between prioritizing arrival success and maintaining continuous connectivity with the base station.

Conversely, when the obstacle density was increased to 30\%, there was a higher likelihood of crashes despite sufficient training. Consequently, during testing, we observed a 10\% failure rate in reaching the destination. Additionally, approximately 25\% of the environment locations exhibited unsatisfactory connectivity efficiency, characterized by high path loss values and weak coverage, particularly in the environment's borders, especially noticeable in the high-frequency bands.

This revision provides a concise summary of the evaluation outcomes and highlights the trade-off between delivery success and maintaining connectivity with the base station during flight operations.

\section{Conclusions}\label{concl}
This paper presented a novel approach to optimizing search and rescue UAV connectivity in challenging terrain using multi Q-learning. We trained two agents: a Strategic Planning Agent for efficient path planning and collision awareness and a Real-time Adaptive Agent to maintain optimal connectivity with a ground base station.

The evaluation results demonstrated the effectiveness of our approach. The Strategic Planning Agent achieved a high success rate in navigating environments with varying obstacle densities. The Real-time Adaptive Agent learned to adapt to different frequency bands, exhibiting faster learning with lower frequency bands due to stronger signals around the environment. We observed a trade-off between arrival success and maintaining continuous connectivity during evaluation and testing. While lower frequencies ensured higher arrival rates, higher frequencies offered more path loss characteristics in far areas.

These findings hold significant promise for real-world search and rescue operations. They highlight the potential of multi Q-learning to enhance UAV autonomy and decision-making in complex environments. Future work could explore incorporating real-time weather data and dynamic obstacle detection for further optimization. Additionally, investigating multi agent coordination for collaborative search patterns could be a valuable advancement.

\section*{Acknowledgment}

This research was funded by EP/X040518/1 EPSRC CHEDDAR and was partly funded by UKRI Grant EP/X039161/1 and MSCA Horizon EU Grant 101086218.

\bibliographystyle{IEEEtran}
\bibliography{IEEEabrv,Bibliography}

\begin{thebibliography}{10}
\providecommand{\url}[1]{#1}
\csname url@samestyle\endcsname
\providecommand{\newblock}{\relax}
\providecommand{\bibinfo}[2]{#2}
\providecommand{\BIBentrySTDinterwordspacing}{\spaceskip=0pt\relax}
\providecommand{\BIBentryALTinterwordstretchfactor}{4}
\providecommand{\BIBentryALTinterwordspacing}{\spaceskip=\fontdimen2\font plus
\BIBentryALTinterwordstretchfactor\fontdimen3\font minus \fontdimen4\font\relax}
\providecommand{\BIBforeignlanguage}[2]{{%
\expandafter\ifx\csname l@#1\endcsname\relax
\typeout{** WARNING: IEEEtran.bst: No hyphenation pattern has been}%
\typeout{** loaded for the language `#1'. Using the pattern for}%
\typeout{** the default language instead.}%
\else
\language=\csname l@#1\endcsname
\fi
#2}}
\providecommand{\BIBdecl}{\relax}
\BIBdecl

\bibitem{san2018intelligent}
V.~San~Juan, M.~Santos, J.~M. And{\'u}jar \emph{et~al.}, ``Intelligent uav map generation and discrete path planning for search and rescue operations,'' \emph{Complexity}, vol. 2018, 2018.

\bibitem{qazzaz2024non}
M.~M. Qazzaz, S.~A. Zaidi, D.~McLernon, A.~M. Hayajneh, A.~Salama, and S.~A. Aldalahmeh, ``Non-terrestrial uav clients for beyond 5g networks: A comprehensive survey,'' \emph{Ad Hoc Networks}, p. 103440, 2024.

\bibitem{salama2023flcc}
A.~Salama, S.~A. Zaidi, D.~McLernon, and M.~M. Qazzaz, ``Flcc: Efficient distributed federated learning on iomt over csma/ca,'' in \emph{2023 IEEE 97th Vehicular Technology Conference (VTC2023-Spring)}.\hskip 1em plus 0.5em minus 0.4em\relax IEEE, 2023, pp. 1--6.

\bibitem{oh2021smart}
D.~Oh and J.~Han, ``Smart search system of autonomous flight uavs for disaster rescue,'' \emph{Sensors}, vol.~21, no.~20, p. 6810, 2021.

\bibitem{zhang2023novel}
C.~Zhang, W.~Zhou, W.~Qin, and W.~Tang, ``A novel uav path planning approach: Heuristic crossing search and rescue optimization algorithm,'' \emph{Expert Systems with Applications}, vol. 215, p. 119243, 2023.

\bibitem{kumar2022obstacle}
G.~Kumar, A.~Anwar, A.~Dikshit, A.~Poddar, U.~Soni, and W.~K. Song, ``Obstacle avoidance for a swarm of unmanned aerial vehicles operating on particle swarm optimization: A swarm intelligence approach for search and rescue missions,'' \emph{Journal of the Brazilian Society of Mechanical Sciences and Engineering}, vol.~44, no.~2, p.~56, 2022.

\bibitem{du2023multi}
Y.~Du \emph{et~al.}, ``Multi-uav search and rescue with enhanced a algorithm path planning in 3d environment,'' \emph{International Journal of Aerospace Engineering}, vol. 2023, 2023.

\bibitem{qazzaz2023low}
M.~M. Qazzaz, S.~A. Zaidi, D.~McLernon, A.~Salama, and A.~A. Al-Hameed, ``Low complexity online rl enabled uav trajectory planning considering connectivity and obstacle avoidance constraints,'' in \emph{2023 IEEE International Black Sea Conference on Communications and Networking (BlackSeaCom)}.\hskip 1em plus 0.5em minus 0.4em\relax IEEE, 2023, pp. 82--89.

\bibitem{hayat2020multi}
S.~Hayat, E.~Yanmaz, C.~Bettstetter, and T.~X. Brown, ``Multi-objective drone path planning for search and rescue with quality-of-service requirements,'' \emph{Autonomous Robots}, vol.~44, no.~7, pp. 1183--1198, 2020.

\bibitem{9508372}
S.~Lins, K.~V. Cardoso, C.~B. Both, L.~Mendes, J.~F. De~Rezende, A.~Silveira, N.~Linder, and A.~Klautau, ``Artificial intelligence for enhanced mobility and 5g connectivity in uav-based critical missions,'' \emph{IEEE Access}, vol.~9, pp. 111\,792--111\,801, 2021.

\bibitem{9836167}
R.~Zahínos, H.~Abaunza, J.~I. Murillo, M.~A. Trujillo, and A.~Viguria, ``Cooperative multi-uav system for surveillance and search\&rescue operations over a mobile 5g node,'' in \emph{2022 International Conference on Unmanned Aircraft Systems (ICUAS)}, 2022, pp. 1016--1024.

\bibitem{qazzaz2024machine}
M.~M. Qazzaz, {\L}.~Ku{\l}acz, A.~Kliks, S.~A. Zaidi, M.~Dryjanski, and D.~McLernon, ``Machine learning-based xapp for dynamic resource allocation in o-ran networks,'' \emph{arXiv preprint arXiv:2401.07643}, 2024.

\bibitem{8794345}
A.~A. Meera, M.~Popović, A.~Millane, and R.~Siegwart, ``Obstacle-aware adaptive informative path planning for uav-based target search,'' in \emph{2019 International Conference on Robotics and Automation (ICRA)}, 2019, pp. 718--724.

\bibitem{9220149}
J.~P. Queralta, J.~Taipalmaa, B.~Can~Pullinen, V.~K. Sarker, T.~Nguyen~Gia, H.~Tenhunen, M.~Gabbouj, J.~Raitoharju, and T.~Westerlund, ``Collaborative multi-robot search and rescue: Planning, coordination, perception, and active vision,'' \emph{IEEE Access}, vol.~8, pp. 191\,617--191\,643, 2020.

\bibitem{sutton2018reinforcement}
R.~S. Sutton and A.~G. Barto, \emph{Reinforcement learning: An introduction}.\hskip 1em plus 0.5em minus 0.4em\relax MIT press, 2018.

\bibitem{munoz2019deep}
G.~Mu{\~n}oz, C.~Barrado, E.~{\c{C}}etin, and E.~Salami, ``Deep reinforcement learning for drone delivery,'' \emph{Drones}, vol.~3, no.~3, p.~72, 2019.

\bibitem{dalela2012tuning}
C.~Dalela, M.~Prasad, P.~Dalela \emph{et~al.}, ``Tuning of cost-231 hata model for radio wave propagation predictions,'' \emph{Academy \& Industry Research Collaboration Center}, 2012.

\end{thebibliography}

\end{document}